\documentclass{article}

     \usepackage[preprint]{neurips_2019}

\usepackage[utf8]{inputenc} 
\usepackage[T1]{fontenc}    
\usepackage{hyperref}       
\usepackage{url}            
\usepackage{booktabs}       
\usepackage{amsfonts}       
\usepackage{nicefrac}       
\usepackage{microtype}      
\usepackage{tikz}
\usetikzlibrary{bayesnet}

\title{Machine Learning for a Low-cost\\Air Pollution Network}

\author{%
  Michael T. Smith \\
  Department of Computer Science\\
  University of Sheffield, UK\\
  \texttt{m.t.smith@sheffield.ac.uk} \\
  \And
  Joel Ssematimba \\
  Department of Computer Science \\
  Makerere University, Uganda \\  
  \texttt{semajoel@gmail.com} \\
  \And
  Mauricio A. \'{A}lvarez \\
  Department of Computer Science\\
  University of Sheffield, UK\\
  \texttt{mauricio.alvarez@sheffield.ac.uk} \\
  \And
  Engineer Bainomugisha \\
  Department of Computer Science \\
  Makerere University, Uganda \\  
  \texttt{baino@cis.mak.ac.ug} \\
}
\begin{document}

\maketitle

\begin{abstract}
Data collection in economically constrained countries often necessitates using approximate and biased measurements due to the low-cost of the sensors used. This leads to potentially invalid predictions and poor policies or decision making. This is especially an issue if methods from resource-rich regions are applied without handling these additional constraints. In this paper we show, through the use of an air pollution network example, how using probabilistic machine learning can mitigate some of the technical constraints. Specifically we experiment with modelling the calibration for individual sensors as either distributions or Gaussian processes over time, and discuss the wider issues around the decision process.
\end{abstract}

\section{Introduction}

We consider the example of a deployment of an air pollution monitoring network in Kampala, an East African city. Air pollution contributes to over three million deaths globally each year\citep{lelieveld2015contribution}. Kampala has one of the highest concentrations of fine particulate matter (PM 2.5) of any African city \cite{mead_2017}. Unfortunately, there is no programme for monitoring air pollution in the city due to the high cost of the equipment required. Hence we know little about its distribution or extent. Lower cost devices do exist, but these do not, on their own, provide the accuracy required for decision makers. In our case study, the Kampala network of sensors consists largely of low cost optical particle counters (OPCs) that give estimates of the PM2.5 particulate concentration. These are known to experience bias depending on humidity \citep{badura2018evaluation} and degrade relatively quickly due to dust and clogging. This network of sensors will soon be supplemented with three reference instruments (certified by MCERTS or equivalent). It is useful to briefly consider the additional issues in the Kampala network, compared to (for example) the LAQN in London. (1) The low-cost OPCs increasingly overestimate PM2.5 in humid conditions. (2) There is considerably more dust (coarse particulates) in the environment in Kampala, leading to sensor degradation. (3) In Kampala pollution exists from additional sources (e.g. road-surfaces, cooking, rubbish-burning, diesel generators, etc). (4) The PM2.5 estimate provided by the OPC is based on assumptions around particle size distributions which are likely to be inaccurate in Kampala.
Regular calibration is clearly necessary, but the sensors typically can't be regularly moved. Thus we will be performing in-situ calibration using a set of mobile sensors installed on motor-bike taxis. This is a similar concept to that described in \citet{kizel2018node}, in which sensors are calibrated in a chain. That model becomes somewhat intractable as the network becomes more complex, and fails to account for the time since calibration. Closely working with the Kampala Capital City Authority (KCCA) we have identified a series of specific requirements for the model output. To summarise: the model should allow a prediction to be made at any location and should attempt to quantify its uncertainty. We chose to use Gaussian process regression (GPR) as this allows us to specify strong priors around the expected spatio-temporal structure of pollution and also provides the necessary uncertainty quantification. Our approach is to assume the low-cost OPC sensors merely measure a scaling to the true pollution value $f$ by a weight $w_i$ specific for that time and sensor $i$. So a given measurement $y$ is given by, $y(t) = w_i(t) f(t) + \epsilon$.
\begin{figure}[t!]
  \centering    
    \includegraphics[width=0.7\columnwidth]{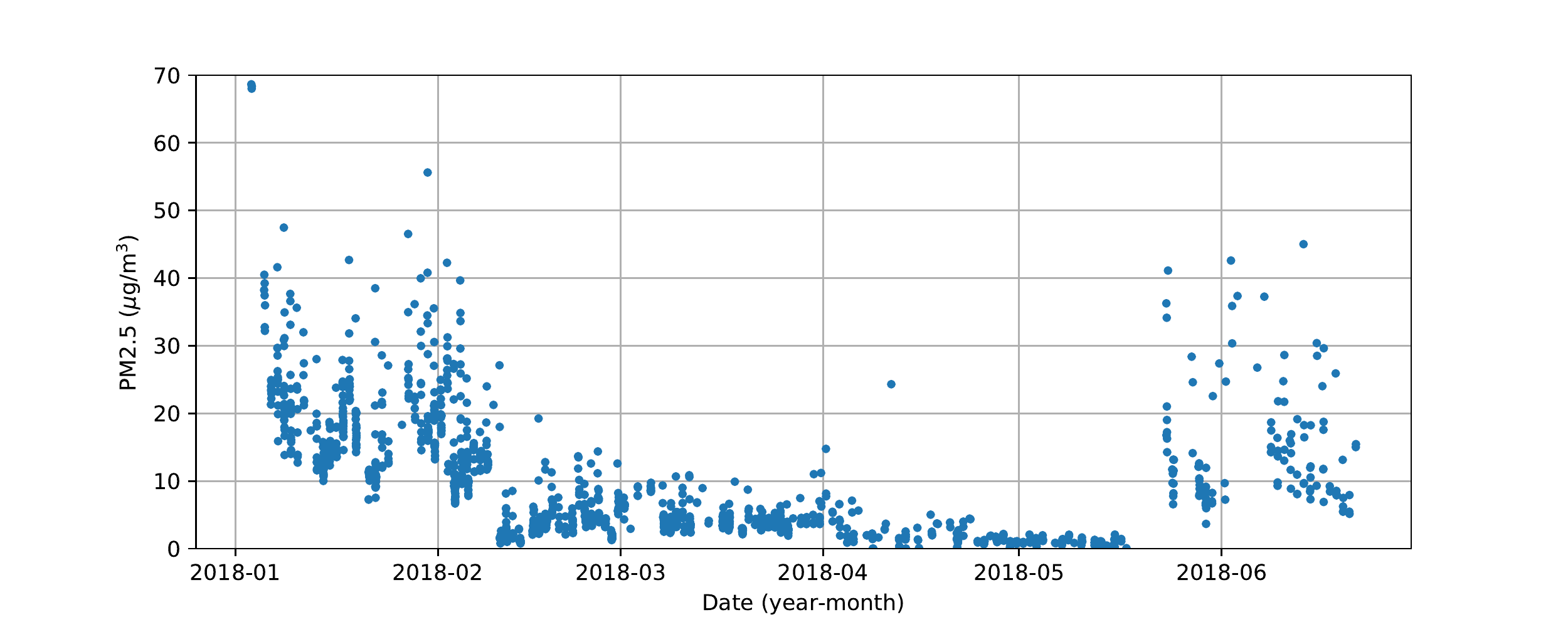}
    \includegraphics[width=0.2\columnwidth]{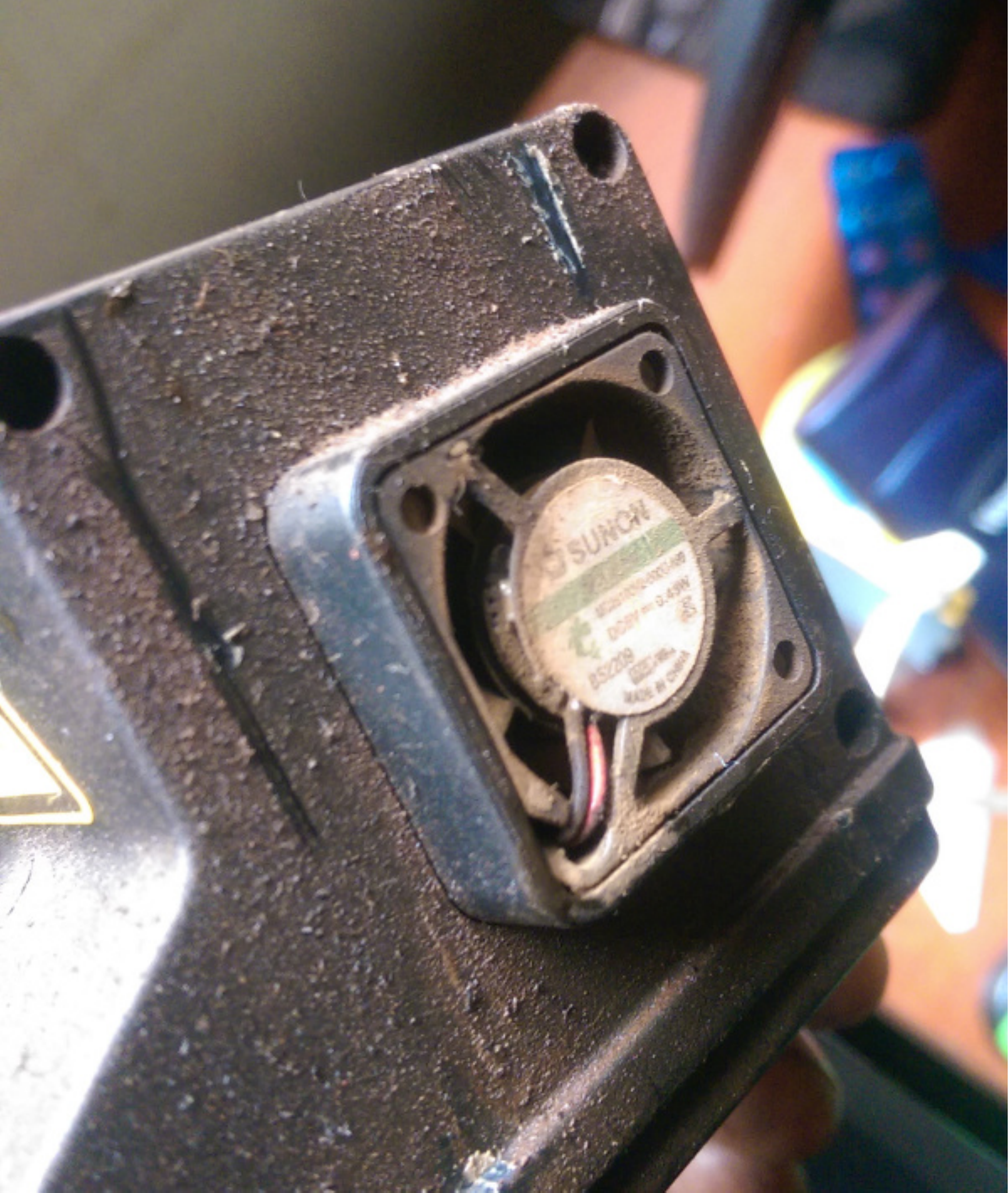}       
  \caption{\emph{Left}, effect of clogging on measured particulate pollution. No co-located reference data was available, but ambient pollution was known to remain mainly in the 10-40 $\mu$g/m${}^3$ range over the whole period. Note how measured value reduces due to dust entering the sensor. Maintenance was conducted in May, leading to a recovery of measurements. \emph{Right} photo of dust clogging the fan.}
    \vspace{-3mm}    
  \label{clogging}
\end{figure} 
This calibration is both uncertain and is known to drift over time. For example Figure \ref{clogging} illustrates the effect of clogging which causes gradual degradation of the sensor.
We therefore develop a model to handle the uncertainty in the measurements.

\subsection{Model}

As this model and system is still being developed we use a numerical approach (using Hamiltonian Monte Carlo, HMC) to allow the model to be adjusted quickly. This also leads to greater sustainability (due to reduced expertise required). However the scalability of the current system is an issue so some of the posterior distribution estimation will need to be via a variational approximation in a deployed solution.

First consider several co-located sensors (one reference and several low-cost). A coregionalised Gaussian process regression model would be suitable for modelling this case \citep{alvarez2011computationally}, using a rank-1 coregionalisation matrix $\mathbf{a} \mathbf{a}^\top$, with the reference instrument's weight $a_1=1$ fixed. Standard maximum likelihood (ML) hyperparameter optimisation leads to weights that reflect the calibration adjustment required to correct for the biases in the low-cost sensors. 
%
\begin{figure}[t!]
  \centering
    \includegraphics[width=0.55\columnwidth,trim=0 -1cm -2cm 0]{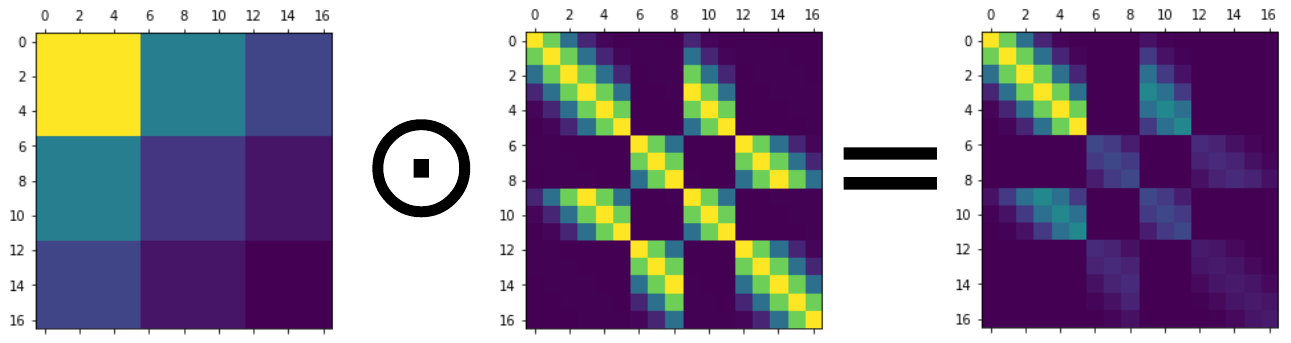}    
  \scalebox{0.6}{
  \tikz{ %
    \node[const] (thetaw) {$\theta_w$} ; %
    \node[latent, right=of thetaw] (z) {GP $z_s^{(j)}(t_s)$} ; %
    \node[latent, right=of z] (w) {$w_i^{(j)}(t)$} ; %
    \node[obs, right=of w] (y) {$y_i^{(j)}(\mathbf{x},t)$} ; %
    \node[latent, above=of y] (f) {GP $f_i(\mathbf{x},t)$} ; %
    \node[const, right=of f] (thetay) {$\theta_y$} ; %
    \plate[inner sep=0.25cm, xshift=-0.12cm, yshift=0.12cm] {plate1} {(w) (y) (f)} {$i=1..N$}; %
    \plate[inner sep=0.25cm, xshift=-0.12cm, yshift=0cm] {plate2} {(z) (w) (y)} {$j=1..M\hspace{5.5cm}$}; %
    \edge {thetaw} {z} ; %
    \edge {z} {w} ; %
    \edge {w,f} {y} ; %
    \edge {thetay} {f} ; %
  }
  }       
  \caption{\emph{Left} Construction of the covariance matrix from three simulated sensors. The coregionalisation structure this produces is in the left most matrix. The second sensor is near the third initially and is then near the first sensor. This leads to the covariance in the central matrix. \emph{Right} The $M$ sensor calibration weights at all observed time points $w_i$ are modelled with a sparse set of pseudo-observations $z_s$ at time points $t_s$. The posterior mean predictions $w_i(t)$ of this sparse GP are used to scale the latent GP $f$'s posterior N predictions to produce the observations $y$. The two GPs hyperparameters are fixed in this model $\theta_w$ and $\theta_y$.}
    \vspace{-3mm}    
  \label{combined}
\end{figure}
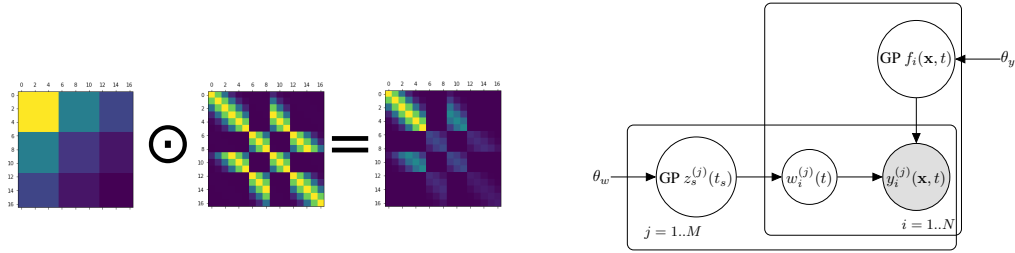
We extend this by considering the case in which the sensors can be in different locations and (some) can move. Modelled with a kernel element-wise product, in which the distance (in space and time) is modelled with a standard EQ kernel, multiplied (element-wise) with the coregionalisation covariance matrix (see figure \ref{combined}). ML optimisation is then able to find calibration values which reflect the sensor biases, even if those sensors are not co-located with the reference instrument - so long as mobile instruments have visited the pairs of instruments. This approach fails however to properly quantify the uncertainty in the calibration, for example due to the changes to the pollution distribution (e.g. road-dust in the dry season). To mitigate this we model the weight hyperparameters $a$ with a series of independent GP priors. For computational efficiency and to aid mixing, we introduced sparsity in these weights, so that, rather than each observation requiring a weight that also is a random variable from a GP, we use a series of latent virtual time points, $t_s$, and observations $z$, and use these to produce the posterior mean vectors $\mathbf{w}^{(j)}$ which is then used to scale the predictions from the GP at the observation locations. Figure \ref{combined} illustrates this slightly more complex model.
\section{Results}
The results below, unlike the example above, are based on simulated data, as reference instruments to validate the method are yet to be deployed. Implemented with Tensorflow Probability.

\textbf{Two sensor demo (GP prior)}
\begin{figure}[t!]
  \centering    
    \includegraphics[width=0.65\columnwidth,trim=0.5cm 0.5cm 0.5cm 0.5cm]{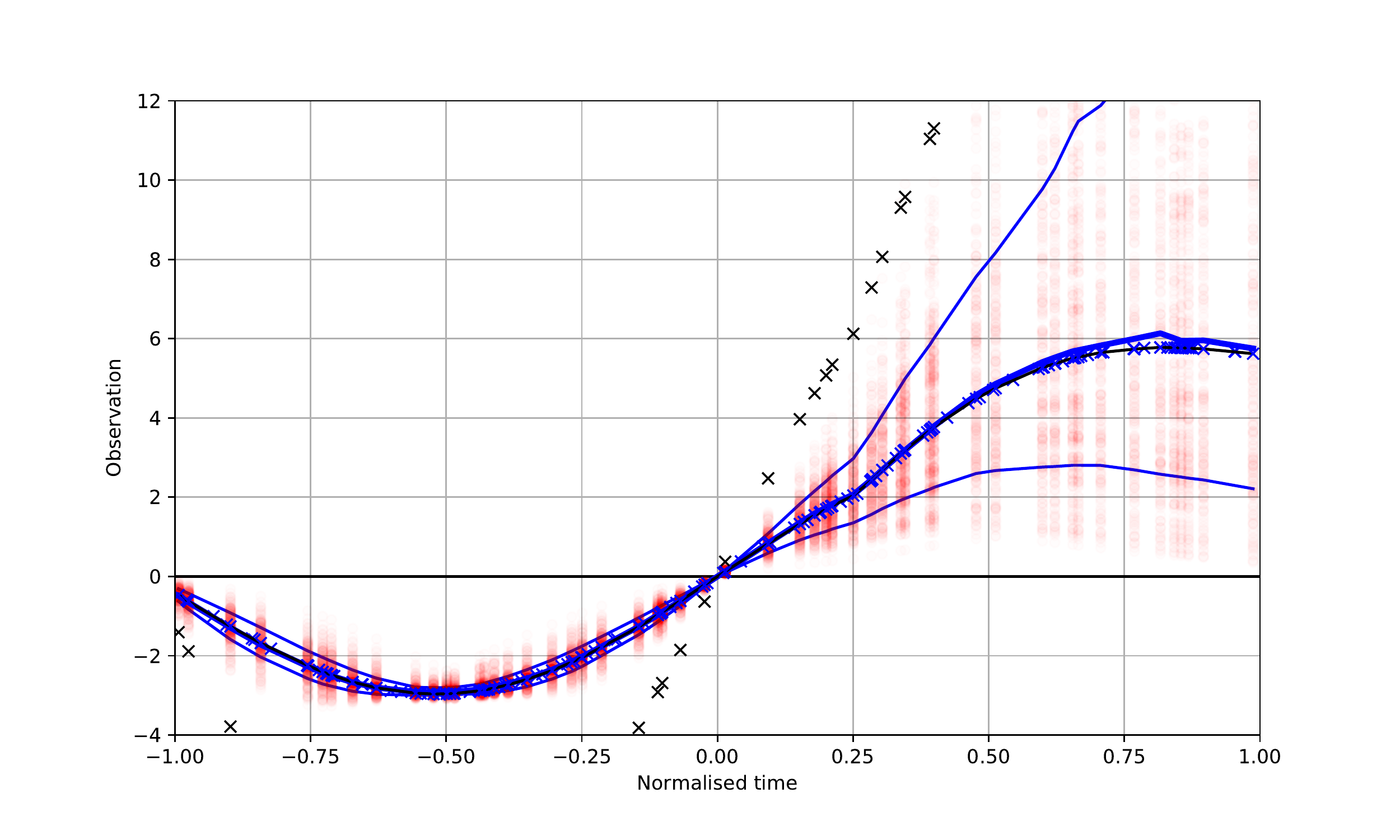}  
    \includegraphics[width=0.3\columnwidth,trim=0.5cm -2cm 0 0]{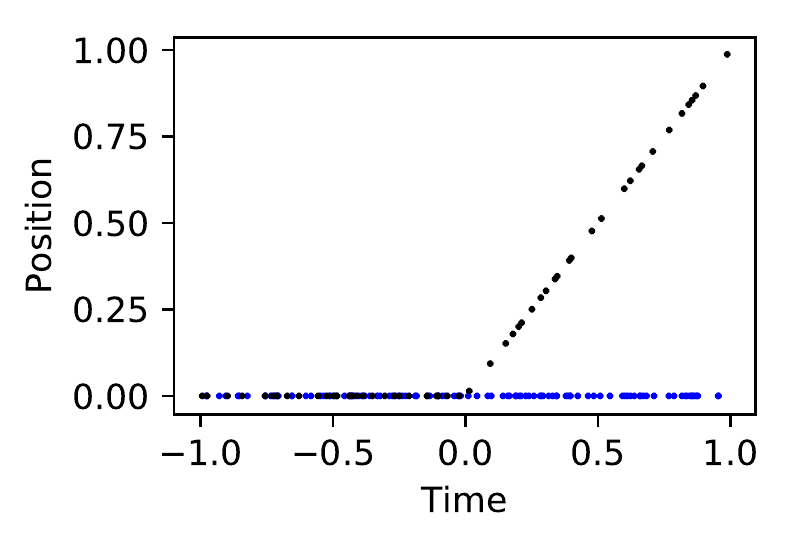}  
  \caption{Left, Simulation data with two instruments. Reference instrument, blue crosses; bias OPC instrument, black crosses ($3 \times$ correct value). For simplicity the correct value at the two instrument locations remains the same (unknown to the model), indicated by black line. Red circles are samples from the MCMC. Thick blue line shows their median and thin blue lines one standard error. Right, spatiotemporal location of both sensors (note the OPC moves away after time zero).}
    \vspace{-3mm}    
  \label{twosensordemo}
\end{figure}
Figure \ref{twosensordemo} shows with the use of a simulation the effect of the GP prior on the scaling weight by considering two simulated sensors (a reference instrument and a low-cost OPC). From time -1 to time zero they are colocated. After time zero the low-cost instrument moves away from the reference. The lengthscale of the scaling GP (exponentiated quadratic kernel) prior has been chosen to be only 4 time units, to demonstrate how the uncertainty in the calibration grows once the reference instrument stops providing support.

\begin{figure}[t!]
  \centering
    \includegraphics[width=0.27\columnwidth,trim=0 0 0 0]{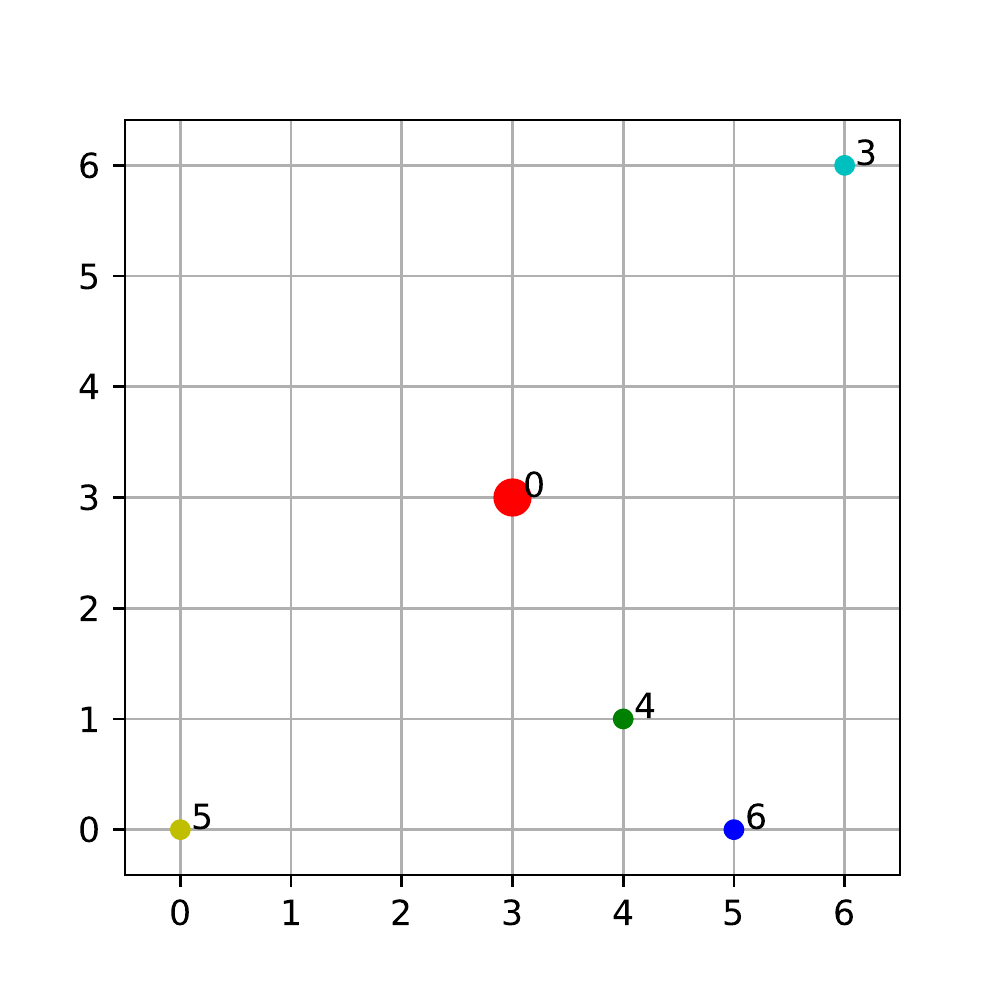}
    \includegraphics[width=0.33\columnwidth,trim=0 0 0 0]{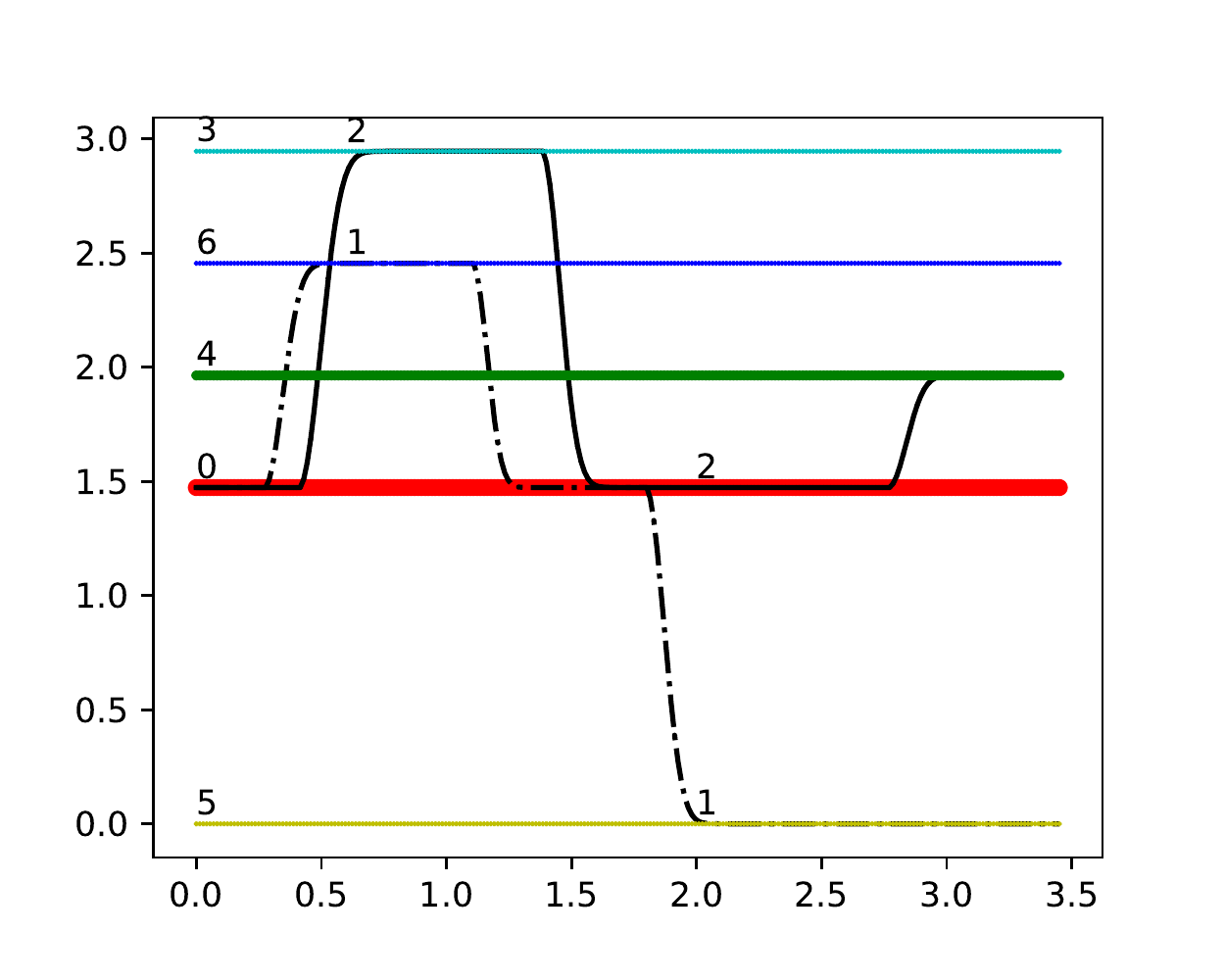}
    \includegraphics[width=1.05\columnwidth,trim=0.5cm 0.5cm 0.5cm 0.5cm]{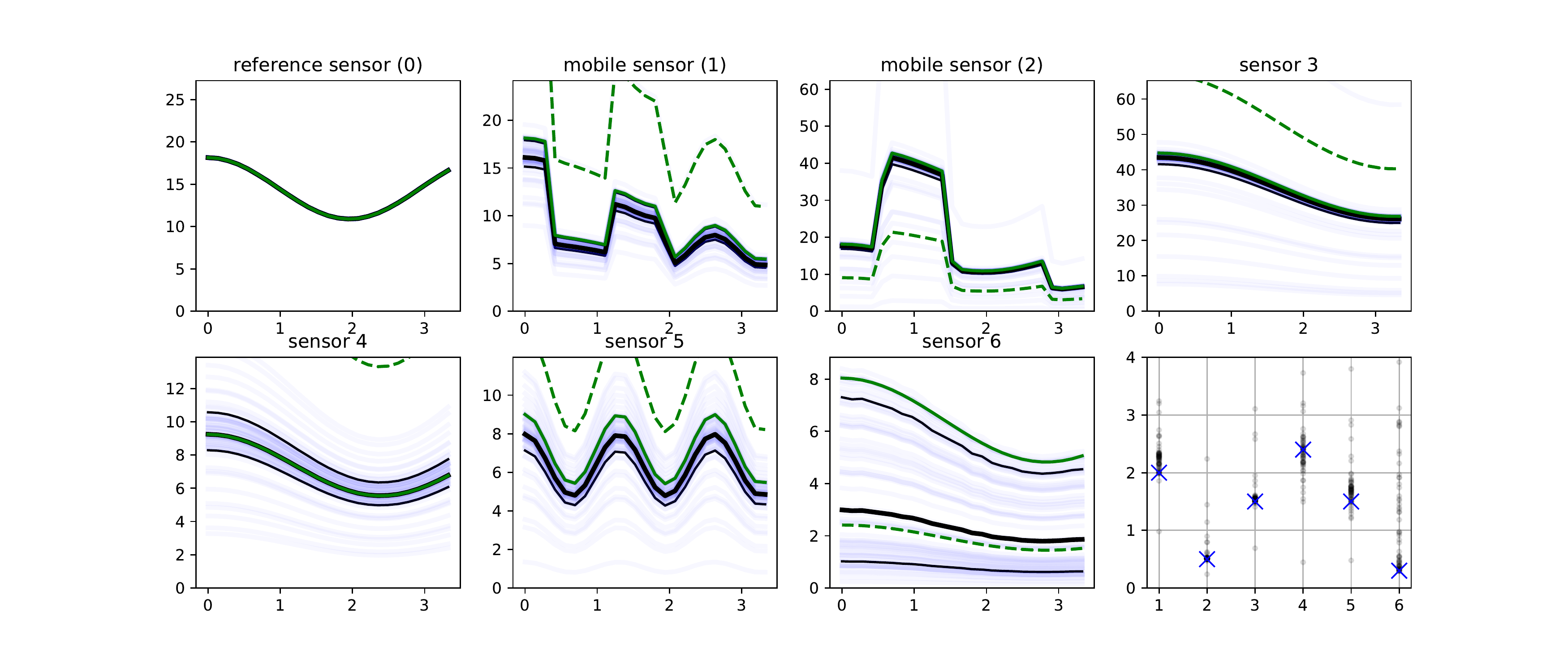}
  \caption{\emph{Upper left}, a map of the four low-cost instruments (labelled 3-6, and the reference instrument 0. \emph{Upper right}, shows how the two mobile sensors (black lines) move between the five static sensors. \emph{Lower plot} Simulation of seven sensors. Measured value indicated with dashed green line. True pollution value indicated with solid green line. Samples from the MCMC based on the measured values indicated with faint blue lines (median, black thick line; standard-error, thin black lines). Figure at lower-right shows MCMC samples of these scaled weights.}
    \vspace{-3mm}    
  \label{simmap}
\end{figure}

\textbf{Sensor network (Gaussian prior)}
We next simulate the more complex situation in which four distant low-cost units need calibrating by using a pair of low-cost mobile sensors to `transport' the calibration signal from the reference instrument. Due to the additional complexity we initially use a simple Gaussian prior on the weights (so the calibration is assumed to be fixed over time).
Figure \ref{simmap} illustrates the locations of the simulated sensors and the route that the mobile sensors make between them. Figure \ref{simmap} shows the results. Most locations are fairly well estimated, with the model selecting the correct calibration. Note that mobile sensor (2) is more precisely characterised than (1). This might be because it spent longer at the reference instrument. The upshot though is that weights for sensors (5) and (6), which are visited by sensor (1), are not very accurately estimated. It could be that the prior's variance needs increasing (currently 25) to reduce its effect, as the data is not very informative.
\section{Discussion}
The described probabilistic modelling approach to this issue appears to be relatively robust (the model worked without any careful parameter selection) and straight-forward. The main issue is when the calibration is `chained' (not shown) which leads to very large uncertainty towards the end of the chain. This may just reflect the true limitations of a long calibration chain. The method described assumes that the pollution characteristics will not vary: i.e. the bias of a sensor is independent of location. This assumption can be tested by occasional visits with reference instrumentation across the city. Our next steps are to deploy reference instruments and test this method with the motor-bike taxis and low-cost OPCs already in-place.
%
%
Some issues still need to be resolved: (1) Data collection biases. Specifically the focus on ambient (outdoor) air pollution. Gender roles in the society mean this is a gendered issue. We are actively working with partners to incorporate an indoor element to the monitoring. (2) Opportunity cost of implementation. The money and time spent developing and deploying the network may have been better spent on other development issues. However, training and mentoring are central to the project, with the intention that the research group will reach an international research standard. These indirect benefits may even exceed the direct results, for research projects, where supporting tertiary education is a key development outcome. (3) Sustainability. How long will the system last? Long-term cost? Who can maintain it? (4) The mobile sensors are mounted on motorbike taxis. The routes they take could conceivably contain private data. We are developing differential privacy methods to obscure this in the predictions. (5) (Ab)use of these results. Who will use the data? It is conceivable that it might be used as a reason or excuse to constrain an activity, such as a solid-fuel cooking, which a vulnerable group depends on. The alternative is to with-hold or cancel the monitoring (which also may be unethical).

\textbf{Conclusion} Many similar deployments of low-cost sensors exist as part of the move to `smart-cities'. But the poor quality of the data collected limits its use for policy making. In this paper we suggest a method to quantify these uncertainties, allowing predictions to be made to aid policy making and monitor interventions.

\textbf{Acknowledgments} Project funded by Google, USAID: Development Impact Lab and the UK EPSRC.

\bibliographystyle{plainnat}
\bibliography{refs}
\end{document}